\pdfoutput=1

\documentclass[11pt]{article}

\usepackage[]{acl}

\usepackage{times}
\usepackage{latexsym}
\usepackage{todonotes}

\usepackage[T1]{fontenc}

\usepackage[utf8]{inputenc}

\usepackage{microtype}
\usepackage{multirow}
\usepackage{multicol}
\usepackage{graphicx}
\usepackage{makecell}
\usepackage{booktabs}
\usepackage{color}
\usepackage{colortbl}
\usepackage{subfigure}

\title{Are Large Language Models Good Fact Checkers: A Preliminary Study}

\author{Han Cao$^{1,2}$ \and Lingwei Wei$^{1,}\thanks{*Corresponding author.}$ \and Mengyang Chen$^{1,2}$  \and \\ 
 \textbf{Wei Zhou}$^{1}$ \and \textbf{Songlin Hu}$^{1,2}$ \\
$^1$ Institute of Information Engineering, Chinese Academy of Sciences \\
$^2$ School of Cyber Security, University of Chinese Academy of Sciences \\
\texttt{\{caohan, weilingwei, chenmengyang, zhouwei, husonglin\}@iie.ac.cn} \\
}

\begin{document}
\maketitle
\begin{abstract}
Recently, Large Language Models (LLMs) have drawn significant attention due to their outstanding reasoning capabilities and extensive knowledge repository, positioning them as superior in handling various natural language processing tasks compared to other language models.
In this paper, we present a preliminary investigation into the potential of LLMs in fact-checking. This study aims to comprehensively evaluate various LLMs in tackling specific fact-checking subtasks, systematically evaluating their capabilities, and conducting a comparative analysis of their performance against pre-trained and state-of-the-art low-parameter models. Experiments demonstrate that LLMs achieve competitive performance compared to other small models in most scenarios. However, they encounter challenges in effectively handling  Chinese fact verification and the entirety of the fact-checking pipeline due to language inconsistencies and hallucinations. 
These findings underscore the need for further exploration and research to enhance the proficiency of LLMs as reliable fact-checkers, unveiling the potential capability of LLMs and the possible challenges in fact-checking tasks.
\end{abstract}

\section{Introduction}

Fact-checking is a pivotal task that aims to assess the verdict of a check-worthy claim with the retrieved evidence, which has drawn significant attention in research fields \cite{DBLP:journals/tacl/GuoSV22, DBLP:conf/coling/ThorneV18}. In detail, fact-checking can be divided into four subtasks: check-worthiness detection, evidence retrieval, fact verification, and explanation generation. To solve these tasks, a mass of automatic fact-checking frameworks and models have been proposed \cite{DBLP:conf/acl/KimPKJTC23, DBLP:conf/acl/WangHCASM23, savchev2022ai, DBLP:conf/acl/LuHZMC23}. In detail, they utilize pre-trained language models to extract textual features of both claim and evidence and leverage Deep learning methods to fuse and obtain comprehensive representations to predict the verdict of the claim.

However, due to the limitation of the scale of training data, the aforementioned methods are domain-sensitive, which cannot be easily used to fact-check claims that belong to other domains without re-training the model. For instance, the model targeted at politics cannot be directly used to verify the financial claim \cite{rangapur2023finfact}. Besides, it requires a large amount of annotated claims to train these methods, which is time-consuming and laborious. 

With the significant advances in Large Language Models (LLMs) and their great capability of understanding and reasoning human language, LLMs, such as GPT-4 \cite{openai2023gpt4}, and LLaMa2 \cite{touvron2023llama}, have been widely utilized to deal with several Natural Language Processing (NLP) tasks \cite{bai2023audiolog, vanveen2023clinical, taffa2023leveraging}, and have shown excellent performance which outperforms the state-of-the-art low-parametric models. Some attempts to utilize LLMs to figure out fact verification, one of the subtasks of fact-checking, have been already made and shown the potential capability of LLMs for fact-checking \cite{DBLP:journals/peerj-cs/ZengZ22, DBLP:conf/acl/ZengG23, lee2021towards}. However, there are no systematic evaluations of the performance of LLMs on all of these subtasks of fact-checking. Hence, to demonstrate the ability of LLMs to fact-check claims and offer feasible and valuable future research direction, we aim to systematically evaluate the whole process of fact-checking, and analyze the potential usefulness of LLMs to fact-checking tasks.

In this paper, we investigate the following Research questions (RQs) to evaluate whether LLMs are good fact-checkers:
\begin{itemize}
    \item \textbf{RQ1: }In the 0-shot setting, with different prompts, can LLMs perform excellently on fact-checking tasks?
    \item \textbf{RQ2: }For the lack of ensemble models that can deal with the fact-checking pipeline, can LLMs solve all of the subtasks simultaneously?
    \item \textbf{RQ3: }Is prompt tuning useful to improve the ability of LLMs to fact-check?
    \item \textbf{RQ4: }For the large-scale training corpora and knowledge, can LLMs utilize their knowledge as a knowledge base to offer evidence to a claim that requires to be verified?
\end{itemize}

To figure out these RQs, we conduct several experiments on 3 fact-checking datasets and elucidate the performance of LLMs on each subtask and the whole pipeline, respectively. The experimental results indicate that it is hard for LLMs to solve fact-checking tasks in the 0-shot settings even though we use different methods to enhance the prompts (\textbf{RQ1}). Besides, LLMs can fact-check claims and give reasonable explanations simultaneously, but fail to come up with the performance when LLMs only solve the fact verification task (\textbf{RQ2}). As for other tasks, prompt tuning can still work on fact-checking tasks (\textbf{RQ3}). They can flexibly use their knowledge to retrieve relevant evidence to predict the label as well. Nevertheless, there is still a long way to go to eliminate hallucination and false information when LLMs are utilized as a knowledge base (\textbf{RQ4}). The experimental results show that more attempts and research are required to improve the performance of LLMs on fact-checking tasks and to make them good fact-checkers.

\section{Related Work}
In this section, we will first introduce the development of language models from the perspective of pre-trained models and large language models in section \ref{lm}. Then, we will talk about the fact-checking task and its subtasks in detail in section \ref{fc}

\subsection{Language Model}
\label{lm}
\subsubsection{Pre-trained Model}
Pre-trained Models (PTMs) are models pre-trained with a large-scale corpus and able to solve the downstream NLP tasks. Inspired by the Transformer \cite{vaswani2017attention} framework, several PTMs have been proposed in succession, such as BERT \cite{devlin2019bert}, RoBERTa \cite{liu2019roberta}, and DeBERTa \cite{he2021deberta}. It has demonstrated that PTMs can outstandingly solve most NLP tasks.

\begin{itemize}
    \item BERT is a pre-trained model that leverages a bidirectional Transformer framework and only takes advantage of the encoder part, which enhances the capability of language understanding.
    \item RoBERTa is an extension of the BERT model that utilizes large-scale unsupervised data and advanced training techniques to achieve state-of-the-art results on various NLP tasks.
    \item DeBERTa is an advanced natural language processing model designed to capture fine-grained linguistic relationships. It improves upon the BERT architecture by employing disentangled attention mechanisms and enhanced decoding strategies.
\end{itemize}

\begin{figure*}[t]
    \centering 
    \includegraphics[width=0.94\linewidth]{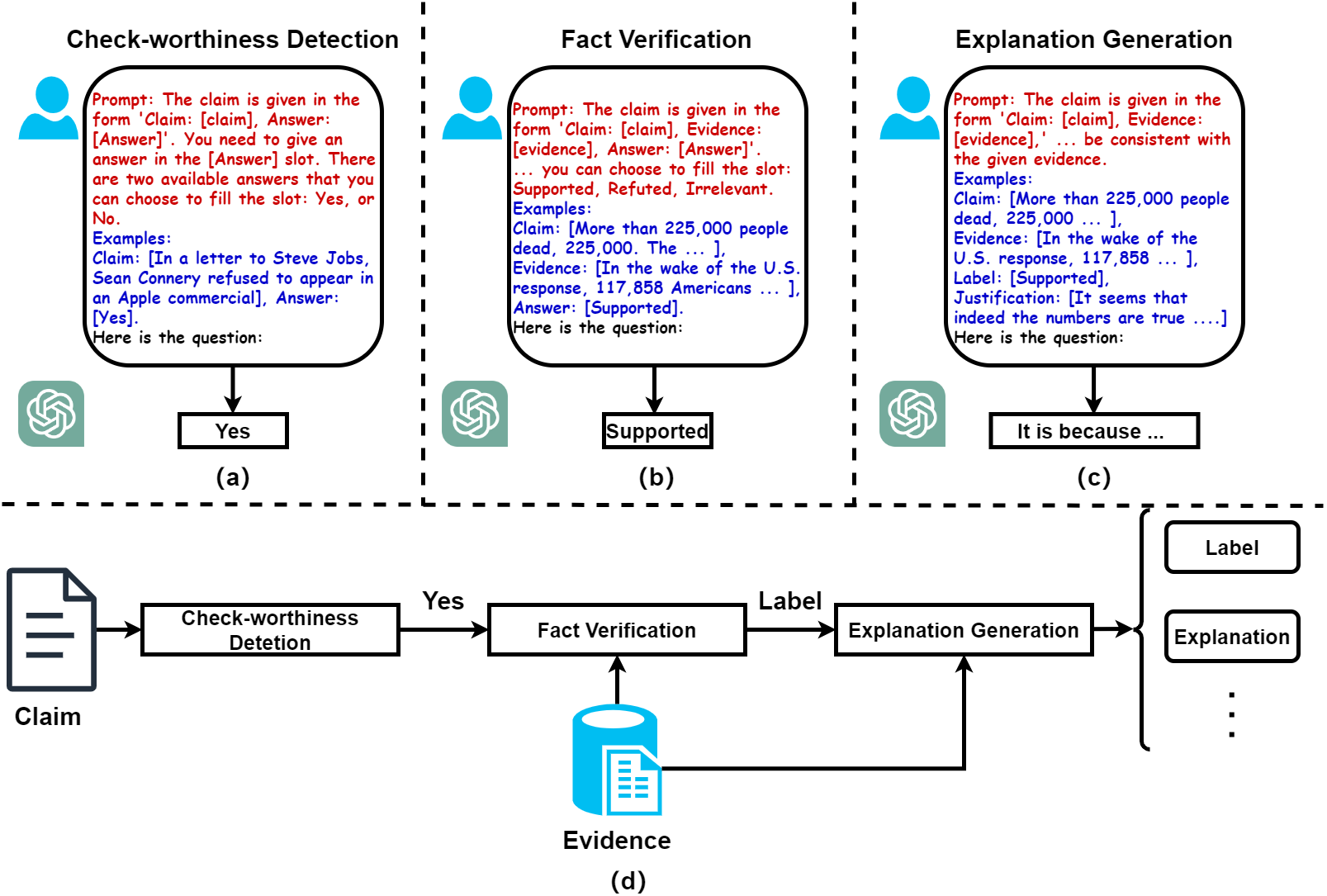} 
    \caption{Framework of the evaluation. Figure (a), (b), and (c) are frameworks of our experiments in Check-worthiness Detection, Fact Verification, and Explanation Generation tasks respectively. Figure (d) denotes the pipeline Fact-Checking task in detail.} 
    \label{Fig-frame} 
\end{figure*}

\subsubsection{Large Language Model}
With the rapid development of PTMs, researchers are interested in whether the scale of the pre-training dataset and the number of parameters can affect the performance of PTMs.Therefore, Large Language Models (LLMs) are proposed, such as ChatGPT, GPT-4 \cite{openai2023gpt4}, ChatGLM \cite{du2022glm}, and LLaMa2 \cite{touvron2023llama}, which contain a larger number of parameters than PTMs. Owing to the large scale of the pre-training dataset, LLMs are proven to be flexible enough to solve cross-domain tasks \cite{nasir2023llms}. Moreover, based on the Prompt Tuning \cite{liu2023gpt} and LoRA \cite{hu2021lora} techniques, LLMs can be fine-tuned to fit the downstream tasks by only using less than 20 samples. 

\begin{itemize}
    \item ChatGPT is a language model developed based on the GPT-3.5 architecture. It's designed for generating human-like text and engaging in conversational interactions. 
    \item GPT-4 is an advanced language model based on the GPT-3 architecture with a larger scale of training data. It incorporates state-of-the-art techniques to further improve natural language understanding and has the best performance on both unimodal and multimodal NLP tasks.
    \item ChatGLM is an open bilingual (English \& Chinese) bidirectional dense model jointly developed by Tsinghua University and Zhipu AI, with 130 billion parameters, pre-trained using the algorithm of the General Language Model (GLM).
    \item LLaMa2 is an advanced model based on LLaMa developed by Meta Organization. It is trained with a larger size of the pre-training corpus and with the technique of Rejection Sampling and Proximal Policy Optimization to attain the best ability.
\end{itemize}

\subsection{Fact-Checking}
\label{fc}
Fact-checking aims to assess the verdict of a check-worthy claim with the retrieved evidence \cite{DBLP:journals/tacl/GuoSV22, DBLP:conf/coling/ThorneV18, DBLP:journals/llc/ZengAZ21}. As shown in Figure~\ref{Fig-frame}, the whole pipeline of fact-checking can be further divided into 4 subtasks: (1) Check-worthiness Detection, (2) Evidence Retrieval, (3) Fact Verification, and (4) Explanation Generation. Recently, numerous methods have been proposed to solve these tasks and achieve better performance.

\paragraph{Check-worthiness Detection.} Check-worthiness detection is a task that aims to check whether a statement is objective, unharmful, and contains information that is worthy of fact-checking. Recent methods take advantage of pre-trained language models and other deep learning mechanisms to extract features and make predictions \cite{savchev2022ai, DBLP:conf/clef/BuligaR22, DBLP:conf/clef/AgrestiHC22, DBLP:conf/clef/EyubogluASK22}. 

\begin{figure*}[t]
    \centering 
    \includegraphics[width=\linewidth]{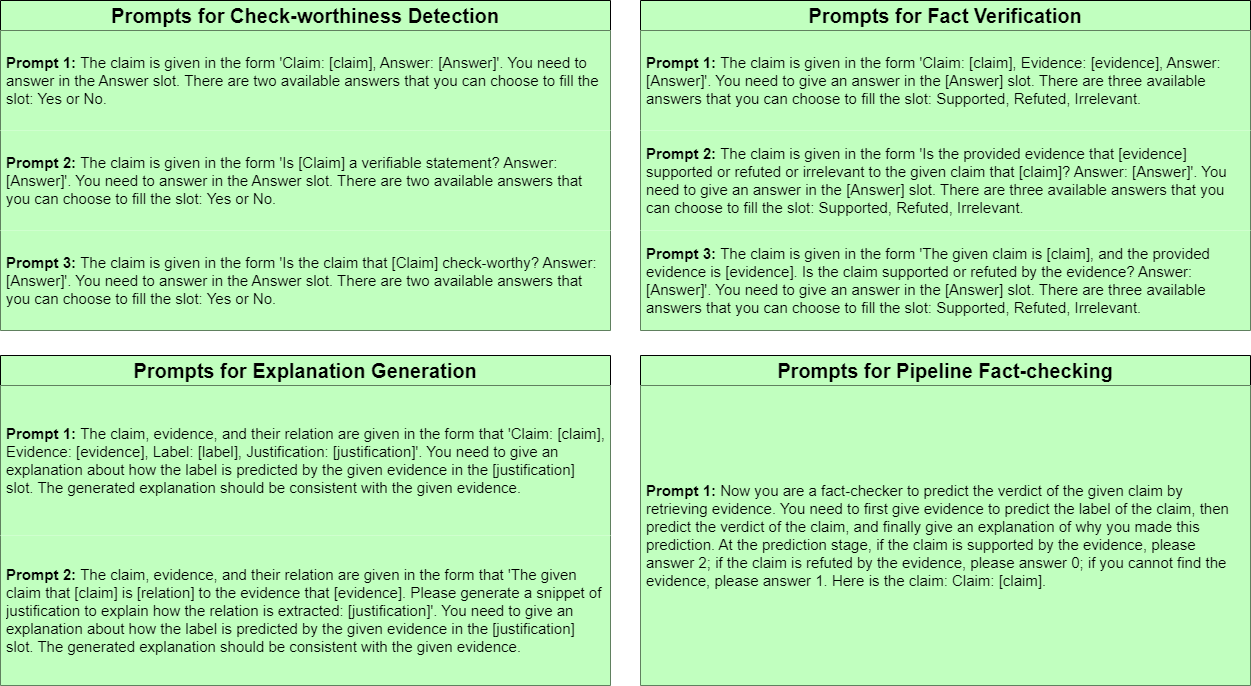} 
    \caption{Prompts designed for each task.} 
    \label{Fig-appd-1} 
\end{figure*}

\paragraph{Evidence Retrieval.} Evidence retrieval targets retrieving relevant evidence that helps to verify the label of the claim. With various methods of Information Retrieval, the proposed frameworks retrieve articles or documents from the network and use the distance or similarity score between claims and evidence to select the most relevant evidence \cite{DBLP:conf/sigir/ChenZGFC22, DBLP:journals/corr/abs-2310-11675, DBLP:conf/emnlp/ShiZYL21, DBLP:conf/naacl/SamarinasHL21}. 

\paragraph{Fact Verification.} Fact verification focuses on predicting the verdict of the claim according to its retrieved evidence. According to the retrieved evidence, fact verification models extract claim and evidence features respectively and fuse them to learn comprehensive understanding and representation to predict the verdict of the given claim \cite{DBLP:conf/acl/KimPKJTC23, DBLP:conf/acl/WangHCASM23, DBLP:conf/emnlp/00600PSC21, DBLP:conf/emnlp/Aly022}. 

\paragraph{Explanation Generation.} Explanation generation is a task that attempts to generate the process of thought and how the prediction is made. There are two common ways to generate explanations for fact-checking. The first one is to draw a summarization according to the evidence and to indicate the procedure of how the prediction is made \cite{DBLP:conf/acl/FajcikMS23, DBLP:conf/acl/PanWLLWKN23}. Different from the former, the second way tries to filter the irrelevant sentences of evidence and gather the rest of the evidence as an explanation \cite{DBLP:conf/icassp/YangVSR22, DBLP:conf/sigir/HuHGWY23, DBLP:conf/aaai/SiZZ23}. Compared to the second method, it is more challenging and complicated for the model to generate a snippet of reasonable and fluent explanations.

\section{Empirical Study}

\subsection{Evaluation Methods}
To evaluate the performance of LLMs on fact-checking tasks, we utilize LLMs to solve these subtasks and the whole pipeline respectively. We first design and select the proper prompts for each task both manually and automatically. Then, we pre-process the data to fit the prompts and use them to conduct these experiments. Figure \ref{Fig-frame} shows the framework of the whole evaluation process.

We first utilize LLMs (we choose ChatGPT\footnote{https://chat.openai.com/}) to generate 10 prompts and templates that they can understand for each task. We randomly select 10 examples of each task and test each candidate. Taking both performance and diversity, we finally choose 3 candidates for check-worthiness detection, fact verification task, 2 for explanation generation task, and 1 for the whole pipeline experiment. Figure \ref{Fig-appd-1} shows the prompts we use for each task.

It has proved to be useful to improve the performance of LLMs to prompt tuning them by giving several examples within the prompt \cite{liu2023gpt}. To further analyze the impact of prompts on LLMs, we use the prompt tuning mechanism with 0-shot, 1-shot, and 3-shot settings. 

Besides, to investigate the performance of different prompts, we use the Chain-of-Thought (CoT) technique \cite{DBLP:conf/nips/Wei0SBIXCLZ22} and task definition to enhance these prompts respectively. For the CoT-enhanced prompts, we follow \citet{DBLP:conf/nips/Wei0SBIXCLZ22} to add \textit{Think step by step} at the end of each prompt. For the task-definition-enhanced prompts, we describe the definition and objective of these tasks in detail with the description in Wikipedia\footnote{https://en.wikipedia.org/wiki/Fact-checking}.

\subsection{Experimental Setups}

\paragraph{Datasets}
To evaluate the performance of LLMs on each subtask of fact-checking, we use three validation sets of fact-checking benchmarks to conduct these experiments: \textbf{CheckThat! Lab} \cite{DBLP:conf/clef/NakovBMAMCKZLSM22}, \textbf{AVeriTeC} \cite{schlichtkrull2023averitec}, and \textbf{CHEF} \cite{DBLP:conf/naacl/HuGWLWY22}. CheckThat! Lab is a dataset for the check-worthiness detection task consisting of 195 English tweets. AVeriTeC and CHEF are the English and Chinese fact verification datasets respectively. Besides, we use AVeriTec for the explanation generation task, for it has ground truth along with the claim and evidence.

\paragraph{Evaluation Metrics}
For the check-worthiness detection, we follow \citet{DBLP:conf/clef/NakovBMAMCKZLSM22} that the F1 score of the Positive class is used for evaluation. For the English fact verification, we follow \citet{schlichtkrull2023averitec} and utilize Accuracy to evaluate the performance of each model. For the Chinese fact verification, we follow \citet{DBLP:conf/naacl/HuGWLWY22} and use Micro- and Macro-F1 scores for the evaluation. For the explanation generation, Meteor Score \cite{banerjee2005meteor} and BLEU-4 \cite{papineni2002bleu} are used to evaluate the quality of the generated explanation.

\paragraph{Implementational Details}
The maximum length of the input token is set to 2048 for GLM-6b, 3000 for GLM-130b, and 4096 for LLaMa2-7b. The batch size is set to 1 for all of these LLMs. We use a Tesla V100-PCIE GPU with 32GB memory for all the experiments.

\begin{figure}[t] 
    \centering 
    \includegraphics[width=\columnwidth]{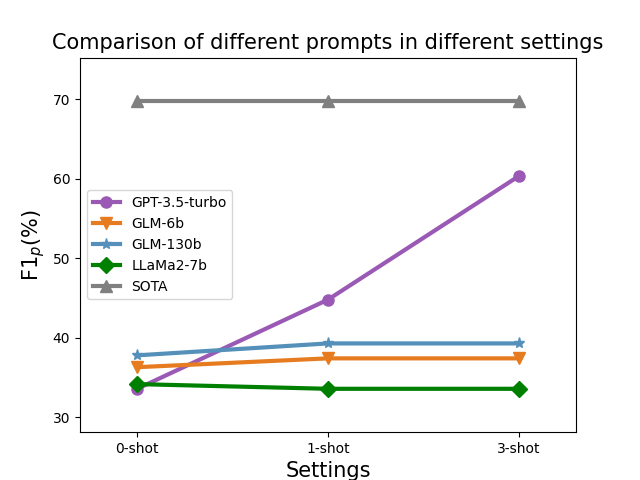} 
    \caption{Comparison of the performance of different LLMs in different settings on check-worthiness detection task. The SOTA result is followed \citet{savchev2022ai}.} 
    \label{Fig-0} 
\end{figure}

\subsection{Performance on Check-worthiness Detection (RQ1 \& RQ3)}

We evaluate the performance of GPT-3.5-turbo, GLM-6b, GLM-130b, and LLaMa2-7b on check-worthiness detection. The results are shown in Figure \ref{Fig-0}. It demonstrates that GPT-3.5-turbo gets the best performance in 1-shot and 3-shot settings, but there is still a huge gap between LLMs and the SOTA model \cite{savchev2022ai}. It is shown that all of these LLMs have worse performance when there are no given examples, mainly because this dataset contains timely information that is out of the range of the training set of LLMs. Along with the increase in the number of examples, the performance of LLMs gets better, especially for GPT, which indicates that the prompt tuning technique can improve the performance of all LLMs if the prompt is perfectly designed and examples are chosen deliberately.

\begin{figure*}[t] 
    \centering
	\subfigure[Confusion matrix of 1-shot setting.]
    {\begin{minipage}{0.49\linewidth}
		\centering
		\includegraphics[width=0.9\linewidth]{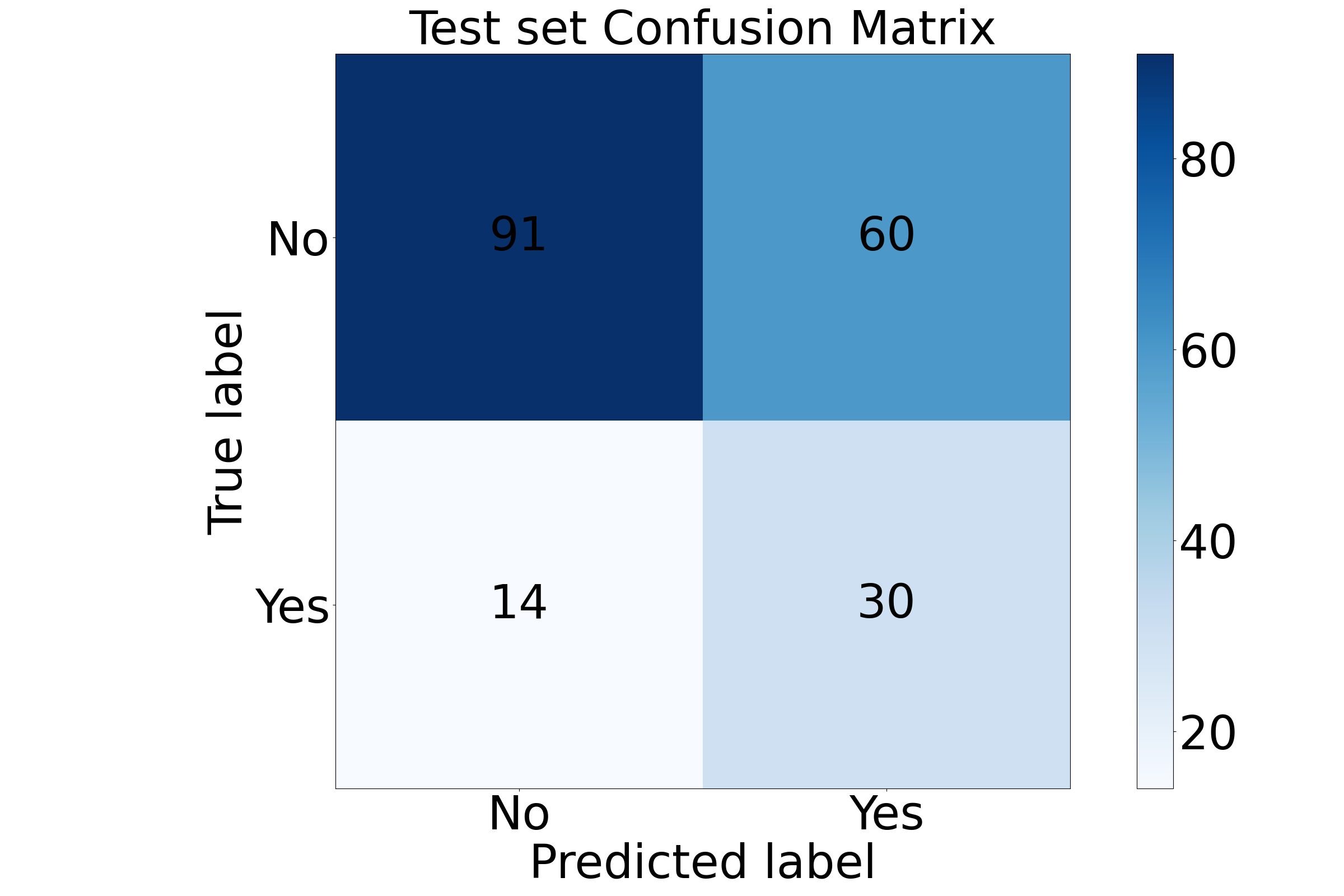}
		\label{F-4-1}
	\end{minipage}}
    \subfigure[Confusion matrix of 3-shot setting.]
	{\begin{minipage}{0.49\linewidth}
		\centering
		\includegraphics[width=0.9\linewidth]{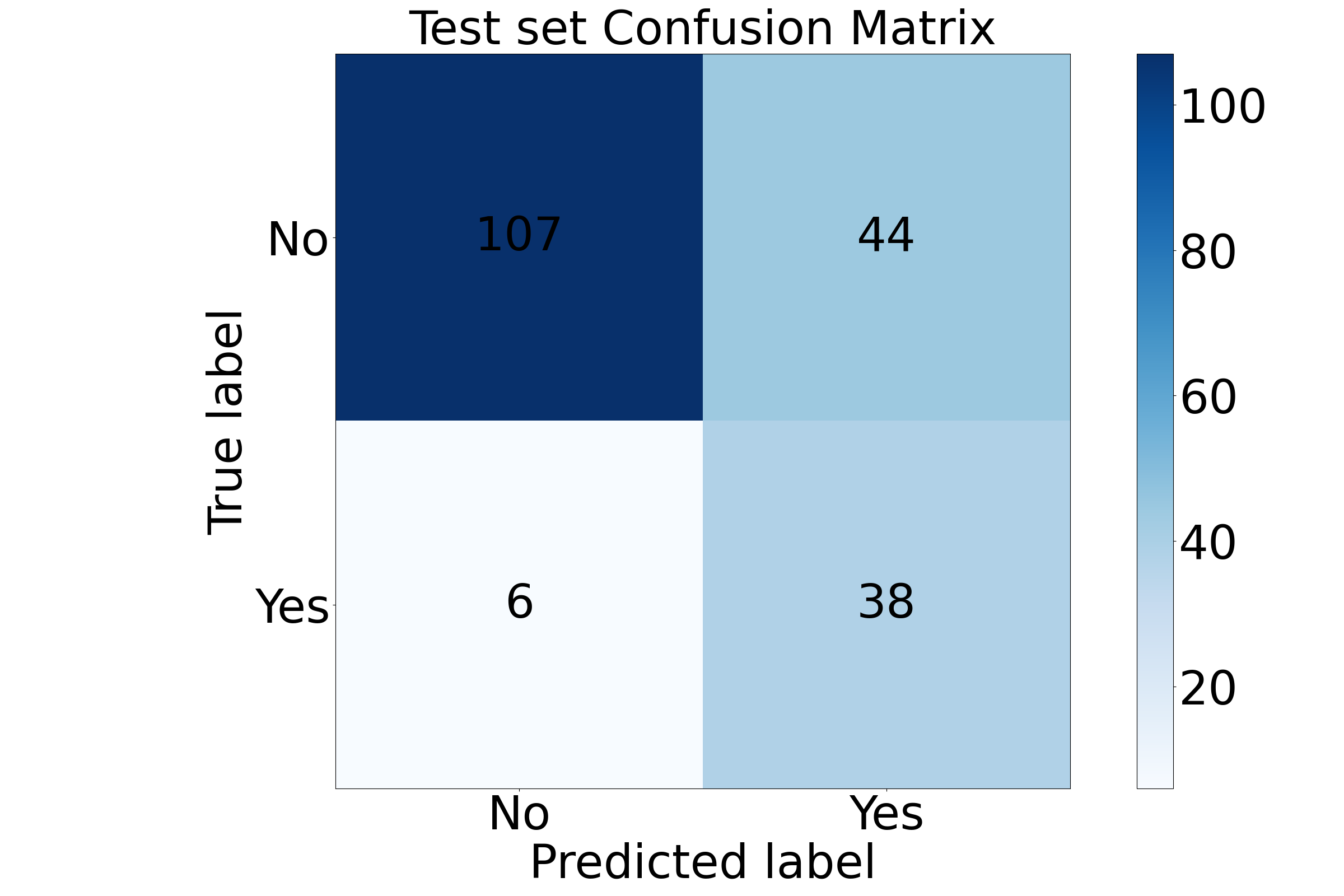}
		\label{F-4-2}
	\end{minipage}}
    \caption{Confusion matrices of different settings on check-worthiness detection task.}
    \label{F-4}
\end{figure*}

We further compare the performance of different LLMs in the 0-shot setting with different prompt enhancements. Table \ref{table-1} shows the results. It can be observed that for GPT and LLaMa2, CoT-enhanced and task-definition-enhanced prompts can improve their performance and make them better understand the objective of this task, whereas for GLMs these enhanced prompts seem not to work well. 

\begin{table}[!tbp] 
    \centering 
    \caption{Comparison of performance of different LLMs on check-worthiness detection in the 0-shot setting. We use the F1 score of the Positive class ($F1_p$) to evaluate the performance. \textbf{Bold} denotes the best performance of each LLM under various prompts. The SOTA result is followed \citet{savchev2022ai}.} 
    \label{table-1} 
    \vspace{5pt} 
    \begin{tabular}{l|c} 
    \hline 
    Model &$F1_p$ (\%)\\
    \hline
    SOTA  & 69.80\\
    \hline
    GPT-3.5-turbo &  \\
    \ \ \ -Standard Prompt & 33.56\\
    \ \ \ -Task Definition & 34.06\\
    \ \ \ -Chain-of-Thought & \textbf{35.56}\\
    \hline
    GLM-6b & \\
    \ \ \ -Standard Prompt & \textbf{36.28}\\
    \ \ \ -Task Definition & 36.12\\
    \ \ \ -Chain-of-Thought   & 35.56\\
    \hline
    GLM-130b & \\
    \ \ \ -Standard Prompt & \textbf{37.77}\\
    \ \ \ -Task Definition & 36.81\\
    \ \ \ -Chain-of-Thought   & 33.48\\
    \hline
    LLaMa2-7b & \\
    \ \ \ -Standard Prompt & 34.15\\
    \ \ \ -Task Definition & 34.50\\
    \ \ \ -Chain-of-Thought   & \textbf{35.07}\\
    \hline
    \end{tabular}
\end{table}

\paragraph{Error Analysis}
Figure \ref{F-4} shows the confusion matrix of GPT-3.5-turbo in the 1-shot and 3-shot settings. With the increase in the given examples, GPT learns to classify check-worthy claims more clearly and improve its capability. It is shown in Figure \ref{F-4-2} that in the 3-shot setting, GPT-3.5-turbo has a high recall score and a comparatively low precision score for the positive class. It elucidates that GPT-3.5-turbo is capable of correctly classifying the check-worthy claims, but the exact boundary still remains ambiguous and needs to be further investigated.

\subsection{Performance on Fact Verification (RQ1 \& RQ3)}
\label{sec-fc}
\begin{figure}[t] 
    \centering 
    \includegraphics[width=\columnwidth]{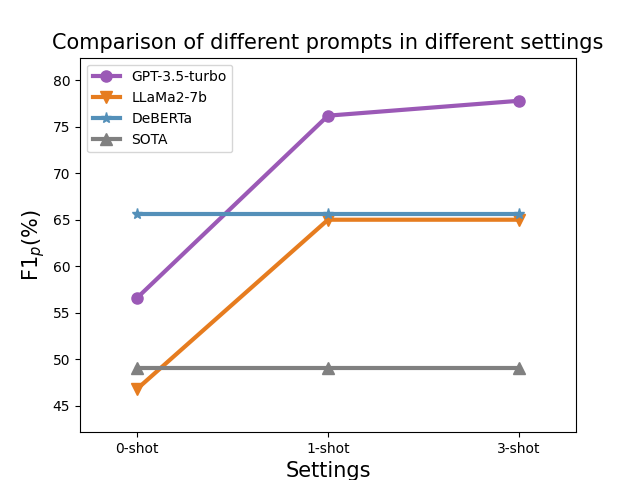} 
    \caption{Comparison of the performance of small models and different LLMs in few-shot settings on English fact verification task. The SOTA result is followed \citet{schlichtkrull2023averitec}.} 
    \label{Fig-5} 
\end{figure}

In the fact verification task, we evaluate the performance of GPT-3.5-turbo and LLaMa2-7b in English and Chinese, for the limitations of the number of tokens of the other two models\footnote{GLM-6b and GLM-130b are not evaluated because the number of tokens hinders the evaluation once we input the evidence together with the claim.}. The experimental results are shown in Figure \ref{Fig-5}, Table \ref{table-2} and Table \ref{table-3}, respectively.

\begin{figure*}[t] 
    \centering 
    \subfigure[Confusion matrix of 1-shot setting.]
    {\begin{minipage}{0.49\linewidth}
		\centering
		\includegraphics[width=0.9\linewidth]{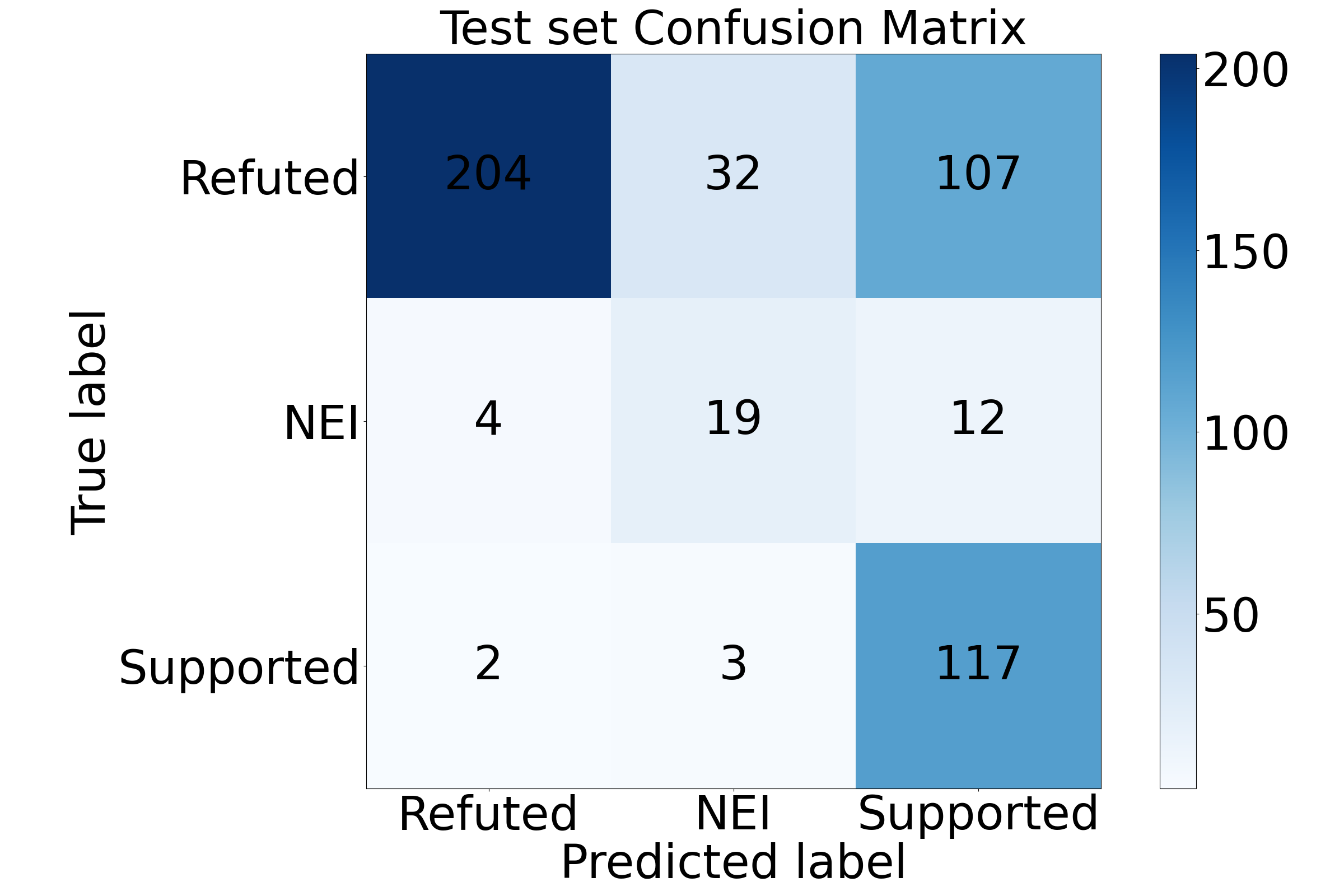}
		\label{F-6-1}
	\end{minipage}}
    \subfigure[Confusion matrix of 3-shot setting.]
	{\begin{minipage}{0.49\linewidth}
		\centering
		\includegraphics[width=0.9\linewidth]{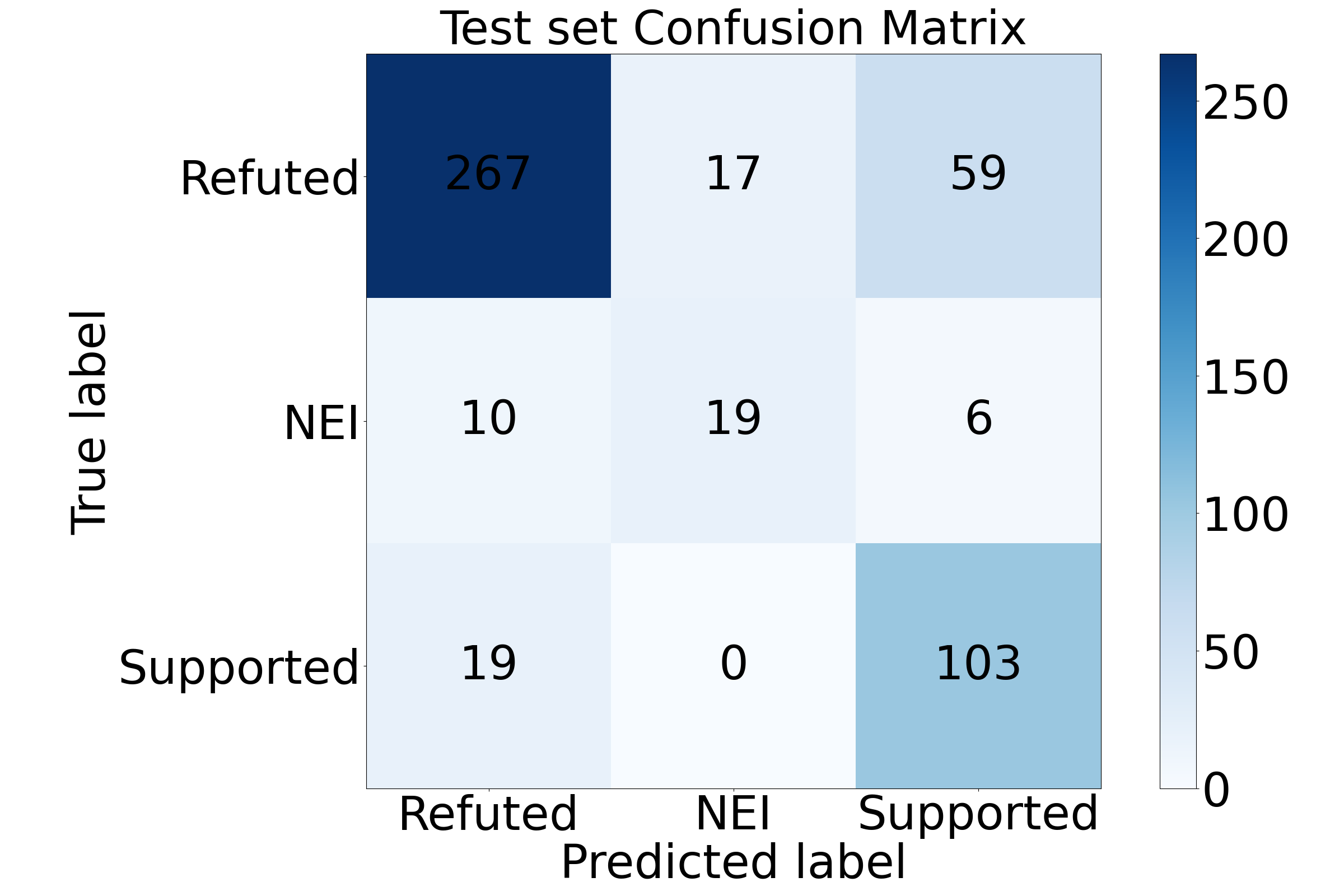}
		\label{F-6-2}
	\end{minipage}}
    \caption{Confusion matrices of different settings on English fact verification task.}
    \label{F-6}
\end{figure*}

\begin{table}[!tbp] 
    \centering 
    \caption{Comparison of performance of different LLMs on English fact verification in 0-shot setting. We use the Accuracy (Acc) to evaluate the performance. \textbf{Bold} denotes the best performance of each LLM. The baseline model is proposed in \citet{schlichtkrull2023averitec}.} 
    \label{table-2} 
    \vspace{5pt} 
    \begin{tabular}{l|c} 
    \hline 
    Model &  Acc (\%)\\
    \hline
    Baseline & 49.00\\
    \hline
    DeBERTa & 65.60\\
    \hline
    GPT-3.5-turbo & -\\
    \ \ \ -Standard Prompt & 56.60\\
    \ \ \ -Task Definition & 63.10\\
    \ \ \ -Chain-of-Thought & \textbf{72.40}\\
    \hline
    LLaMa2-7b & -\\
    \ \ \ -Standard Prompt & 46.80\\
    \ \ \ -Task Definition & 49.30\\
    \ \ \ -Chain-of-Thought & \textbf{57.90}\\
    \hline
    \end{tabular}
\end{table}

\begin{table}[t]
    \centering 
    \caption{Comparison of performance of different LLMs on Chinese fact verification. We use the Micro F1 ($F1_{mic}$) and Macro F1 ($F1_{mac}$) to evaluate the performance. \textbf{Bold} denotes the best performance of each LLM. The SOTA result is followed \citet{DBLP:conf/naacl/HuGWLWY22}.} 
    \label{table-3} 
    \vspace{5pt} 
    \resizebox{\linewidth}{1.0in}{
    \begin{tabular}{l|c|c|c} 
    \hline 
    Model & n-shot & $F1_{mic}$ (\%) & $F1_{mac}$ (\%)\\
    \hline
    SOTA & - & 79.94 & 78.47\\
    \hline
    \multirow{3}{*}{\makecell[l]{GPT-3.5-turbo\\ \ \ \ -English prompt}} & 0-shot & \textbf{28.63} & \textbf{26.46}\\
    & 1-shot & 21.72 & 21.93\\
    & 3-shot & 21.92 & 20.47\\
    \hline
    \multirow{3}{*}{\makecell[l]{GPT-3.5-turbo\\ \ \ \ -Chinese prompt}} & 0-shot & 27.93& 23.51\\
    & 1-shot & {31.73} & 22.70\\
    & 3-shot & \textbf{35.14} & \textbf{33.51}\\
    \hline
    \multirow{3}{*}{LLaMa2-7b} & 0-shot & \textbf{31.93} & 27.92\\
    & 1-shot & 29.93 & \textbf{28.58}\\
    & 3-shot & - & - \\
    \hline
    \end{tabular}}
\end{table}

\subsubsection{English Fact Verification} 
\paragraph{Overall Performance} From Figure~\ref{Fig-5}, GPT-3.5-turbo outperforms other models, even better than the baseline model \cite{schlichtkrull2023averitec} and the pre-trained model, demonstrating that LLMs can perform English fact verification tasks. It can be observed that GPT-3.5-turbo is a good learner when it is given some examples in the input context. The improvement is dramatically huge between 0-shot and 1-shot experiments. However, when giving more examples, the performance increases slowly, perhaps it reaches the best capability of GPT-3.5-turbo on fact-checking without any augmentations or enhancements. For LLaMa2-7b, in the 1-shot and 3-shot settings, it can get a higher Accuracy compared to the baseline model. 

Besides, we also use enhanced prompts to investigate whether these techniques are helpful or not. The experimental results are shown in Table \ref{table-2}. Both the CoT-enhanced and task-definition-enhanced prompts can improve the performance of GPT and LLaMa2. This is probably because the process of verifying a claim needs multi-step thinking to finally make a prediction. 

\paragraph{Error Analysis} The confusion matrix of GPT-3.5-turbo in the 1-shot and 3-shot experiments is shown in Figure \ref{F-6}. Compared to the 1-shot setting, GPT can better classify claims that belong to the \textbf{Refuted} category, while it fails to distinguish the \textbf{Supported} claims. Besides, it is obvious that it is capable for GPT-3.5-turbo to classify Supported and Refuted claims in both settings. However, during the experiments, we find that GPT-3.5-turbo has an ambiguous principle to determine whether the evidence is enough to verify the claim. 

\subsubsection{Chinese Fact Verification} 
\paragraph{Overall Performance} The experimental results are demonstrated in Table \ref{table-3}. The performance of all of these LLMs underperforms the SOTA method \cite{DBLP:conf/naacl/HuGWLWY22}, no matter which language is used to construct prompts. Comparatively, utilizing prompts in Chinese performs better than utilizing prompts in English. It is still challenging for GPT-3.5-turbo to deal with Chinese fact-checking. It is perhaps because GPT-3.5-turbo and LLaMa2-7b are mainly trained on English training data, which makes them confused when they are dealing with other languages. 

\paragraph{Language Consistency} After we investigate the lower performance using English prompts, we reckon that the inconsistency of the language of the prompt and data can negatively impact the performance of LLMs. To further prove it, we use the same prompts written in Chinese as the input to evaluate the performance of GPT-3.5-turbo. Interestingly, when the input context is consistent in the same language between prompt and data, the performance will be better. Besides, when we use English prompts to conduct Chinese fact verification tasks, surprisingly, the performance of LLMs decreases. However, when we replace the English prompts with Chinese prompts, the performance gets better along with the increase in the number of the given examples. This indicates that language inconsistency in prompts will impact the understanding capability of LLMs and further degrade their performance. Though GPT performs better with Chinese prompt, there is still a huge gap between it and the SOTA model.

\begin{table}[t] 
    \centering 
    \caption{Comparison of performance of different LLMs on explanation generation. We use the Meteor Score to evaluate the performance. \textbf{Bold} denotes the best performance of each LLM. The baseline model is proposed in \citet{schlichtkrull2023averitec}} 
    \label{table-4} 
    \vspace{5pt} 
    \begin{tabular}{l|c|c} 
    \hline 
    Model & n-shot & Meteor Score\\
    \hline
    Baseline & - & 0.1100\\
    \hline
    \multirow{3}{*}{GPT-3.5-turbo} & 0-shot & 0.3084\\
    & 1-shot & 0.3285\\
    & 3-shot & \textbf{0.3512}\\
    \hline
    \multirow{3}{*}{LLaMa2-7b} & 0-shot & \textbf{0.1569}\\
    & 1-shot & 0.1488\\
    & 3-shot & - \\
    \hline
    \end{tabular}
\end{table}

\begin{table*}[!t] 
    \centering  
    \caption{Comparison of performance of different LLMs on pipeline fact-checking task. We use Accuracy to evaluate the performance of the Fact Verification task, and Meteor Score, BLEU-4, and Cosine Similarity (Cos-Sim) to evaluate the generated evidence. \textbf{Bold} denotes the best result, and \underline{Underline} denotes the second-best result. The baseline model is proposed in \citet{schlichtkrull2023averitec}.} 
    \label{table-5}
    \vspace{5pt} 
    \begin{tabular}{l|c|c|c|c} 
    \hline 
    \multirow{2}{*}{Model} & Fact Verification & \multicolumn{3}{|c}{Evidence Generation} \\
     & Acc (\%) & Meteor Score& BLEU-4& Cos-Sim \\
    \hline
    Baseline &  49.00 & - & - & -\\
    \hline
    DeBERTa &  \textbf{65.60} & - & - & -\\
    \hline
    GPT-3.5-turbo & \underline{51.80}& \textbf{0.1844} & \textbf{0.2390} & \textbf{0.8585}\\
    \hline
    GLM-130b & 26.20 & \underline{0.1836} & \underline{0.2224} & \underline{0.8499}\\ \hline
    LLaMa2-7b & 39.40 & - & - & -\\
    \hline
    \end{tabular}
\end{table*}

\subsection{Performance on Explanation Generation (RQ1 \& RQ3 \& RQ4)}

We evaluate the performance of GPT-3.5-turbo and LLaMa2-7b in explanation generation tasks and the experimental results are shown in Table \ref{table-4}. GPT-3.5-turbo outperforms the baseline model \cite{schlichtkrull2023averitec}. When some examples are given, GPT-3.5-turbo can provide more concise and accurate justifications for the given claim and its relevant evidence. However, LLaMa2-7b gets worse performance, even though some examples are given. Therefore, GPT-3.5-turbo is capable of generating explanations to help human beings understand better the decision-making process and the wrong part of the given claim according to its relevant evidence, which indicates that GPT-3.5-turbo can be further used to help fact-checking models to be interpretable. As for LLaMa2-7b, we find that it is more likely to imitate and repeat the prompt, instead of using its knowledge to generate a reasonable explanation. It is inferred that this results in different training objectives and training methods of GPT-3.5-turbo and LLaMa2-7b.

\begin{figure*}[t]
    \centering 
    \includegraphics[width=0.8\linewidth]{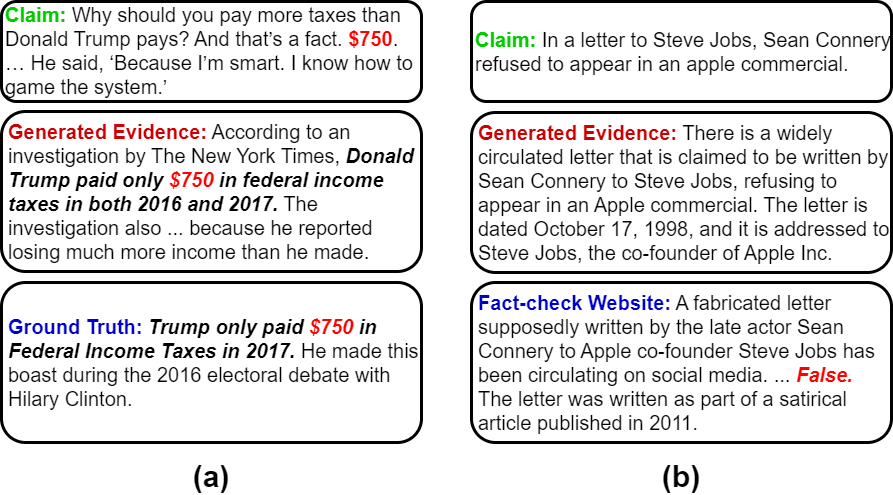} 
    \caption{Examples of generated evidence by GPT-3.5.turbo. Figure (a) shows a claim correctly classified and figure (b) shows a claim wrongly predicted.} 
    \label{Fig-4} 
\end{figure*}

\subsection{Performance on Pipeline Fact-Checking (RQ2)}

Due to the lack of an ensemble model for the whole fact-checking pipeline, we investigate whether LLMs can solve the whole fact-checking pipeline simultaneously or not. Besides, for an automatic fact-checking model, it is often required to generate a complete explanation to interpret how the prediction is made, which makes the model more readable and interpretable. Moreover, along with the advances in LLMs, researchers are questioning whether LLMs can be used as an enormous knowledge base to assist the evidence retrieval task.
Therefore, we use the AVeriTeC dataset to evaluate the performance of GPT-3.5-turbo, GLM-130b, and LLaMa2-7b on the pipeline fact-checking task, containing evidence generation, fact verification, and explanation generation tasks, in the 0-shot setting. The experimental results are shown in Table \ref{table-5}. 

\subsubsection{Fact Verification}
GPT-3.5-turbo outperforms other LLMs and the baseline model. It demonstrates that LLMs can successfully predict the label of claims only taking the claim as input and they can search for relevant evidence automatically to help them make predictions. However, compared with the results shown in Table \ref{table-2}, LLMs have worse performance without the gold evidence. It is supposed that, without the help of retrieved evidence, LLMs fail to better utilize their knowledge to filter noise and collect helpful evidence. Moreover, they may even take misinformation, fake news, or rumors as credible evidence, which is regarded as hallucination (see details in \ref{evi-retrieval}). Thus, LLMs are capable of solving the pipeline fact-checking task simultaneously, whereas the problems of hallucination may impede their performance.

\subsubsection{Evidence Generation}
\label{evi-retrieval}
To further analyze \textbf{RQ4}, we compare the difference between the gold evidence and the evidence generated by LLMs. The results are demonstrated in Table \ref{table-5}\footnote{LLaMa2-7b is not evaluated because it only generates the prediction of the claim, and fails to offer reasonable and meaningful evidence.}.

The generated evidence gets a high Meteor Score and BLEU-4 score. It can be observed that both GPT-3.5-turbo and GLM-130b successfully retrieve relevant and credible evidence for the verification procedure, indicating the fantastic retrieval ability of LLMs. For the explanation generation, compared to the results in Table \ref{table-4}, LLMs outperform the baseline model. The reason why the BLEU-4 scores of explanations generated by LLMs are relatively low is perhaps because of the different language styles. 

Figure \ref{Fig-4}(a) shows an example of the ground truth evidence and the evidence generated by GPT-3.5-turbo. It can be observed that LLMs can generate more complicated and more comprehensive evidence than the ground truth. Besides, they can grab and concentrate on the key phrases that are crucial to help make predictions and search for targeted evidence. 

However, although LLMs show great capability of retrieval, they still have the problem of hallucination, especially factual hallucinations. As shown in Figure \ref{Fig-4}(b), GPT-3.5-turbo regards the news of Sean Connery's letter as true information and uses it to verify the truthfulness of the claim, but this news is proved to be false by REUTERS\footnote{https://www.reuters.com} in November 2020. 

\subsubsection{Discussion}
According to the experimental results of fact verification and evidence generation tasks, we can find that LLMs have the ability to solve the fact-checking pipeline simultaneously. However, the performance is worse than that of the small-sized fine-tuned models. Besides, LLMs can offer and generate solid and convincing evidence to help fact-check the claims, whereas there is still the problem of hallucination that makes LLMs generate false information as evidence, which leads to worse performance.

\section{Conclusion and Future Work}

In this paper, we evaluate the performance of several LLMs on fact-checking tasks. The experimental results demonstrate that LLMs are capable of dealing with these tasks respectively. LLMs can understand the requirements of each task and leverage their knowledge to solve it to some extent. The experiments show that CoT-enhanced and task-definition-enhanced prompts are helpful on some tasks, such as fact verification, but they fail to improve models in other tasks like check-worthiness detection. Besides, LLMs are able to retrieve relevant and meaningful evidence by themselves to verify the truthfulness of claims to some extent, while there are still some problems to be solved, such as hallucination, to improve the performance and effectiveness of LLMs when they are used to deal with pipeline fact-checking tasks.

However, though the performance of LLMs is satisfactory to some extent, there are still some problems that need to be solved to improve their performance. The outputs of LLMs are not always in the form that we need, which is a big challenge to evaluate and analyze the results, especially for classification tasks. Besides, LLMs are better at dealing with English data, but it is hard to handle data in other languages, such as Chinese. 

With the above said, the performance of LLMs on fact-checking still has a great gap compared to that of the models based on other deep learning models, and our work provides the benchmark of using LLMs to solve fact-checking tasks without parametric fine-tuning and gives future research directions to make improvements. In future work, we will contribute to leveraging both the plentiful knowledge held by LLMs and the significant capability of understanding and reasoning to improve the performance of LLMs.

\end{document}